\documentclass[11pt]{article}

\usepackage[preprint]{acl}
\usepackage{times}
\usepackage{latexsym}
\usepackage[T1]{fontenc}
\usepackage[utf8]{inputenc}
\usepackage{microtype}
\usepackage{inconsolata}
\usepackage{amsmath}
\usepackage{amssymb}
\usepackage{booktabs}
\usepackage{graphicx}
\usepackage{multirow}
\usepackage{url}
\usepackage{xcolor}
\usepackage{enumitem}

\newcommand{\bench}{\textsc{ManifestoBench}}

\title{Right Domain, Wrong Code: Benchmarking LLMs on Multilingual Hierarchical Code Assignment}

\author{Junjie Li\thanks{Co-first authors (equal contribution). Authors are listed in reverse alphabetical order by surname.}\thanks{Work done outside Amazon. This work is unrelated to the author's affiliation with Amazon.} \\
  Amazon.com, Inc. \\
  \texttt{junjieli@umich.edu}
  \And
  Jun Fang\footnotemark[1] \\
  University of Michigan \\
  \texttt{junfang@umich.edu}}
\date{}

\makeatletter
\def\ps@arxivfirst{%
  \let\@oddhead\@empty
  \let\@evenhead\@empty
  \def\@oddfoot{\rlap{\footnotesize Accepted to Findings of EACL 2027.}\hfil\thepage\hfil}%
  \let\@evenfoot\@oddfoot
}
\makeatother
\begin{document}
\maketitle
\thispagestyle{arxivfirst}

\begin{abstract}
Can large language models reproduce the fine-grained distinctions of an
established coding scheme? We introduce \bench{}, a multilingual benchmark
derived from the Comparative Manifesto Project (CMP), with 150,000 code-stratified, human-coded
quasi-sentences from 513 election manifestos in 48 countries, 32 document
languages, and 56 codes nested in seven policy domains. We evaluate eight
API models across four families. Four cost-efficient models achieve
21.1--34.3\% exact-code accuracy and 38.4--57.0\% domain accuracy.
On a paired 30,000-instance subset, four stronger models achieve
30.9--48.4\% exact-code accuracy and 53.1--65.4\% domain accuracy.
For the four cost-efficient models, predicting the domain before the code
improves exact accuracy by 0.8--8.6 percentage points; supplying the gold
domain and restricting the candidate set yields 50.2--55.4\% exact accuracy.
Reasoning gains vary by model.
A training-screened ManifestoBERTa reference
reaches 51.9\% on a separate paired subset. Taken together, these results show that broad-domain recognition can mask
substantial error in the final code. This distinction is especially
consequential when the label space is fixed by an external institution:
the model must reproduce the scheme as it stands. \bench{} therefore releases
hierarchy-aware metrics, paired diagnostic conditions, prompts, and
reproducible evaluation code.
\end{abstract}

\section{Introduction}

Large language models (LLMs) are increasingly used as annotators. On familiar
tasks with a small number of intuitive labels, they can approach or exceed
crowd-worker performance \citep{gilardi2023chatgpt,tornberg2025llms} and agree
closely with expert coders \citep{heseltine2024large}. Real annotation systems, however, often rely on coding
schemes that are neither intuitive nor flat. Medical, legal, scientific, and
social-scientific schemes contain related categories arranged in hierarchies,
and their labels encode conventions accumulated by expert communities rather
than ordinary-language semantics alone.

Yet standard LLM benchmarks test broad knowledge, reasoning, or heterogeneous
task collections
\citep{hendrycks2021mmlu,srivastava2022bigbench,liang2023helm}. Such
benchmarks tell us relatively little about whether a model that recognizes a
broad topic can reproduce the fine distinctions in an established coding
scheme. Flat accuracy also hides where failure occurs: a prediction may select
the wrong parent domain, or it may reach the correct domain and choose the
wrong code among neighboring alternatives. Separating these observable
outcomes is therefore necessary before using aggregate accuracy to judge model
capability.

One response to such ambiguity is to adapt the taxonomy itself. Recent
taxonomy-refinement work shows that classification can improve when an LLM
revises a hierarchy to better match its encoded semantics \citep{golde2026taxmorph}.
For coding schemes maintained by external institutions, however, that option
is unavailable: annotators inherit rather than choose the categories, and
merging confusable classes or redefining the label set would change the
prediction target itself. Under this constraint, the model must adapt to the
scheme rather than the scheme to the model.

The Comparative Manifesto Project (CMP) provides a concrete case.%
\footnote{Manifesto Project codings have been reused as shared measurement
infrastructure in social and political sciences research: released code shares and derived scales
support measures of party positions and issue-specific orientations, while
coded quasi-sentences provide filtered corpora for downstream text
analysis
\citep{bohmelt2016party,wratil2023government,malis2024foreign,
garz2024party}. In these applications, CMP
categories are inherited as research inputs rather than redesigned for each study.}
Its trained
human coders assign policy statements from election manifestos to a stable
international coding scheme \citep{werner2021handbook,mpds2025a}. The scheme
distinguishes, for example, incentives for firms from market regulation,
general equality principles from concrete welfare expansion, and economic
growth from underspecified economic goals. Because CMP codes become variables
for comparing parties, countries, and elections, a topically plausible but
incorrect code can change the substantive measurement.

To evaluate this fixed-scheme setting, we introduce \bench{}, a benchmark and
evaluation protocol. The main benchmark is a reproducible, code-stratified
sample of 150,000 quasi-sentence occurrences from 513 manifestos issued
between 2018 and 2025. It covers 32 document languages; English accounts for
only 19.9\% of instances. At this scale, model choice is part of the research
design, so we target corpus-scale annotation under social-science research
budgets. We evaluate eight API models across four provider families: four
cost-efficient models on 150K and all eight on the same paired 30K subset.
We report exact-code accuracy, domain
accuracy, macro and weighted $F_1$, and a decomposition into within-domain,
cross-domain, and invalid errors.

The main benchmark establishes the size of the exact--domain gap. To examine
where that gap arises, we compare three prompt conditions on a fixed
30,000-instance subset: a flat code decision, a domain-then-code decision, and
an oracle condition that supplies the correct domain and lists only its child
codes. The paired design distinguishes errors associated with parent-domain
selection from residual errors among codes within the correct domain. To
separate prompt limitations from task learnability, we also compare the
instruction-only LLMs with ManifestoBERTa on a 51,716-instance subset
screened against its labeled training corpus, providing a task-specific
supervised reference. Paired 30K experiments vary candidate-list
restriction, model capability, reasoning and temperature while preserving
the task and inputs.

Our contributions are:
\begin{enumerate}[leftmargin=*,nosep]
    \item a large multilingual benchmark for LLM annotation under an
    established hierarchical coding scheme;
    \item a hierarchy-aware evaluation protocol that separates exact-code
    reproduction from broad-domain recognition and keeps invalid outputs in
    the evaluation;
    \item a paired diagnostic study of parent-domain and within-domain error,
    together with a training-screened supervised reference; and
    \item supplementary software containing prompts, sampling and
    reconstruction scripts, evaluation code, aggregate results, and audit
    documentation, without redistributed source-linked records.
\end{enumerate}

\section{Related Work}

\paragraph{Political text as data.}
Automated content analysis makes large political corpora tractable, but its
measurements remain task-specific and require validation
\citep{grimmer2013text,barbera2021classification}. Generative LLMs lower the
entry cost further by performing classification without task-specific
training. Studies report strong results for relevance, sentiment, stance, and
ideological positioning \citep{gilardi2023chatgpt,tornberg2025llms,
heseltine2024large,lemens2025positioning}. Broader evaluations nevertheless
find substantial variation across social-science tasks and warn that expert
knowledge and downstream measurement validity cannot be inferred from fluent
outputs \citep{ziems2024css,pangakis2023validation,peskoff2023credible}.
These concerns are especially acute when labels are defined by a codebook
rather than by ordinary-language categories.

\paragraph{Codebook-based annotation.}
In that setting, codebooks operationalize constructs more narrowly than their
ordinary-language
names \citep{halterman2026codebook}, and prompt design can change both compliance and accuracy
\citep{atreja2025prompt}. Most directly, \citet{halterman2026codebook} evaluate
real political-science codebooks, including an English-language Manifesto
Project subset, with open-weight models. They find weak instruction-only
performance and gains from supervised instruction tuning. \bench{} extends
this line with 32 original document languages, 150,000 instances,
eight API models spanning deployment tiers, and a paired hierarchy-aware evaluation that
separates parent-domain errors from errors among codes within the correct
domain.

\paragraph{Automated manifesto classification.}
Manifesto classification includes cross-lingual topic classifiers
\citep{glavas2017crosslingual} and ManifestoBERTa, an XLM-R model
released by the Manifesto Project and fine-tuned on its annotated corpus
\citep{manifestoberta2023}. This supervised work asks how well in-domain
training transfers to held-out text. Our primary question is complementary:
whether general-purpose models can reproduce the coding scheme from
instructions without updating their weights on CMP labels. We nevertheless
include a document-disjoint, exact-text-screened ManifestoBERTa comparison to
anchor the instruction-only scores against task-specific supervision.

\paragraph{Hierarchy and model alignment.}
Hierarchical text classification exploits label relations to address large,
imbalanced label spaces \citep{chalkidis2020lmtc}. Recent methods
enrich labels \citep{paletto2024label}, retrieve hierarchy-aware
demonstrations \citep{zhao2024retrieval}, or use LLMs to refine
taxonomies \citep{golde2026taxmorph}. The last approach treats the
hierarchy as mutable; CMP instead requires stable code identities and
boundaries for longitudinal and cross-national measurement. Structured
prediction also differs from our domain-then-code prompt: separately scored
decisions can be combined under global constraints, whereas sequential
prompting mainly changes the elicitation procedure
\citep{pauk2026structured}. Separately, multilingual benchmarks emphasize per-language
evaluation across broad language coverage
\citep{hu2020xtreme,bandarkar2024belebele}; we therefore evaluate
original-language text and report descriptive language-level results without treating
language groups as causal contrasts.

\paragraph{Benchmark design.}
Together, these strands motivate a fixed, auditable benchmark with explicit
sampling assumptions.
MMLU, BIG-bench, and HELM motivate
standardized comparisons and transparent
metrics \citep{hendrycks2021mmlu,srivastava2022bigbench,liang2023helm}.
Dataset documentation exposes sampling and annotation
assumptions \citep{gebru2021datasheets}. Following these principles, we
preserve fixed samples and an audit trail, report malformed outputs rather
than silently repairing them, and distinguish benchmark scores from population
estimates.

\section{\bench{}}

\subsection{Source Corpus and Coding Scheme}

We combine MPDS 2025a metadata \citep{mpds2025a} with annotated texts from
the Manifesto Corpus \citep{manifestocorpus}. Human coders
segment manifestos into
\emph{quasi-sentences}, each intended to express one political idea, and assign
CMP categories following Coding Instructions Version 5
\citep{werner2021handbook}. We retain 513 manifestos from 48 countries and the
2018--2025 period. Text is evaluated in its original language; we do not
machine-translate it. We identify language once per full manifesto with
\texttt{langid.py} \citep{lui2012langid} and cross-check assignments against
MPDS country metadata. The resulting sample spans 32 languages. Appendix
Table~\ref{tab:language-performance} reports coverage and performance for
each language.

Within this corpus, the task has 56 three-digit main categories nested in
seven domains:
External Relations (1xx), Freedom and Democracy (2xx), Political System
(3xx), Economy (4xx), Welfare and Quality of Life (5xx), Fabric of Society
(6xx), and Social Groups (7xx). Decimal extensions in the corpus, such as
\texttt{606.1}, are mapped to their three-digit main category. We exclude
missing annotations and gold \texttt{000} statements because the benchmark
targets discrimination among substantive main categories. Models may still
output \texttt{000} when they judge an input to lack political content; this is
a validly formatted prediction but is incorrect for this evaluation set.

\subsection{Main 150K Sample}

After normalization, the candidate pool contains 549,968 quasi-sentences with
one of the 56 main codes. A simple random sample would be dominated by frequent
economic and welfare labels. We therefore sample within each three-digit code
using seed 20260629. We target at least 1,000 and at most 5,000 instances per
code, taking all available examples for rarer labels. The resulting benchmark
contains exactly 150,000 occurrences, all 56 labels, all 513 manifestos, and
between 435 and 5,000 examples per label. Consequently, aggregate scores
describe performance on the benchmark distribution, not the natural frequency
distribution of CMP codes.

The sample has 149,213 unique \texttt{manifesto + text} inputs. The difference
comes from identical quasi-sentence strings repeated within a manifesto. The
flat runner sends each unique input once and assigns the resulting prediction
to every corresponding occurrence. There are 590 duplicated-input groups; 161
receive more than one gold label in the corpus. We retain occurrences to respect the
source annotations and report a sensitivity analysis that removes conflicting
groups and collapses unambiguous duplicates.

\subsection{Fixed 30K Comparison Sample}

The 150K sample estimates aggregate performance. To compare prompt conditions
without changing sample composition, we draw a second code-stratified sample
of 30,000 rows from the main benchmark using seed 20260713. It contains all 56 labels, 511 manifestos, no duplicate
sentence identifiers, and complete flat predictions from all eight API models. The
same ordered instances are used for every model and prompt condition. This
paired design prevents condition comparisons from being confounded with sample
composition.

\subsection{Supervised No-Overlap Reference}

The mechanism sample isolates prompt conditions, but it does not show how much
task-specific supervision can learn. To provide that anchor, we also evaluate
the official sentence-only ManifestoBERTa model released for Corpus
2023a \citep{manifestoberta2023}. We query the historical Manifesto Corpus
metadata version \texttt{2023-1} \citep{corpus2023} and exclude every main-sample manifesto
present in that release. The resulting subset contains 54,159
document-disjoint occurrences. We additionally remove 2,443 rows whose
Unicode-normalized text exactly matches a statement in a historical-corpus
manifesto. The resulting screened subset contains 51,716 occurrences from 160
manifestos and retains all 56 gold labels. We run the
frozen 2023 sentence model on the same quasi-sentence text supplied to each
flat-prompt LLM, using its documented 200-token input limit; only three screened
subset instances exceed that limit. This design rules
out document overlap and observed normalized exact-text overlap with
ManifestoBERTa's supervised fine-tuning corpus; it does not claim that public
manifesto text was absent from generic XLM-R pretraining.

\section{Experimental Setup}

\subsection{Models}

We evaluate eight API models, pairing a cost-efficient and a higher-capability
configuration within each of four provider families.
LLM annotation is attractive in social science because it can avoid
task-specific training and reduce the cost of labeling large corpora
\citep{gilardi2023chatgpt,heseltine2024large,ziems2024css}. Political-text
classification has long emphasized scalable but task-validated measurement
\citep{grimmer2013text,barbera2021classification,halterman2026codebook}. Our target is
corpus-scale deployment by social-science researchers: the main benchmark
alone comprises 600,000 model--instance evaluations, making price and
throughput part of the research design. Provider prices differentiate these
deployment tiers \citep{openai2026pricing,google2026pricing,qwen2026pricing,mistral2025mistralsmall,mistral2026medium35}. The model pairs are:
\begin{itemize}[leftmargin=*,nosep]
    \item \textbf{GPT-5.4-nano} and \textbf{GPT-5.6 Sol} through the OpenAI
    Responses API, both with low reasoning effort and low verbosity
    \citep{openai2025gpt54nano,openai2026gpt56sol};
    \item \textbf{Gemini 2.5 Flash Lite}, with a 512-token thinking budget
    \citep{google2025gemini25flashlite}, and \textbf{Gemini 3.8 Flash}, with
    thinking level low, through Google \citep{google2026gemini38flash};
    \item \textbf{Qwen Flash} (\texttt{qwen-flash-2025-07-28})
    and \textbf{Qwen3.8 Max} through Alibaba Model Studio
    \citep{qwen2025qwenflash}, both with thinking enabled and a 256-token budget; and
    \item \textbf{Mistral Small 4} (\texttt{mistral-small-2603}; hereafter
    \textbf{Mistral Small}) \citep{mistral2025mistralsmall} and
    \textbf{Mistral Medium 3.5} \citep{mistral2026medium35} through Mistral's API, both with reasoning
    effort \texttt{none}.
\end{itemize}
Temperature is zero where the provider exposes it. These settings prioritize
a low-reasoning reference within each family. This
choice is also consistent with recent work reporting no uniform benefit from
additional or reward-optimized
reasoning on populism coding and annotator-disagreement modeling
\citep{ni2026reasoning,chalkidis2026populism}.
Appendix~\ref{app:inference-settings} and the software record exact caps and aliases.

All eight models receive the identical flat prompt on 30K. The four
cost-efficient models are also evaluated on 150K and in domain-prompt
experiments and high-reasoning comparisons. Six temperature comparisons vary
0 versus 0.7 at fixed low or high
reasoning for Gemini, Qwen and Mistral; GPT rejects the temperature parameter.
Model and reasoning comparisons use the flat prompt. Temperature comparisons
fix the output cap at 4096 within each reasoning setting.
Appendix~\ref{app:inference-settings} records the configuration for each comparison.

\subsection{Prompt Conditions}

Having fixed the model set, we next standardize elicitation.

\paragraph{Prompt development.}
Before freezing the formal protocol, we explored concise inline instructions,
expanded category descriptions, and attaching the complete CMP handbook as a
provider file. These development runs are not reported as benchmark
conditions. Whole-document attachment did not improve observed pilot behavior
and substantially increased prompt length while introducing provider-specific
file handling. Because prompt structure and answer format can alter model behavior
\citep{atreja2025prompt,halterman2026codebook,labat2026polyglots}, we froze
one concise, provider-independent prompt before the main runs and did not
select among prompt variants on the 150K outcomes. We use
\emph{instruction-only} to mean that no task-specific parameters are
updated; the prompt's illustrative examples are part of the fixed codebook
instruction.

\paragraph{Flat codebook prompt.}
The flat prompt defines the task, lists the 56 canonical
three-digit titles in domain order, gives decision rules and eight short
examples, and requests one code with no explanation. The prompt includes
several distinctions that are frequently confused, such as 401 vs.\ 403 and
503 vs.\ 504. It does not reproduce every paragraph of the CMP handbook.

\paragraph{Hierarchical two-step.}
The model first selects a domain from 0--7 and then a three-digit code from that
domain; 1--7 are the substantive parent domains and 0 denotes no political
content. It returns minimal JSON:
\texttt{\{"step1\_domain": 5, "step2\_code": "504"\}}.
The candidate titles and examples are byte-identical to the flat prompt.

\paragraph{Oracle domain.}
The prompt supplies the gold domain, and its candidate list contains only that
domain's child codes. It retains the specificity and salience rules, key
distinctions, and eight examples.
The model returns one three-digit code. This privileged-information diagnostic
measures code selection after supplying the
parent domain and restricting the candidate set to its legal children. Output
\texttt{000} is parsed as a valid code but recorded as violating the oracle
candidate constraint.

\paragraph{Gold domain/full list.}
This condition supplies the same gold-domain instruction and replaces the
restricted candidate block with all 56 titles. Both arms supply the correct
domain. We compare both conditions on 30K, with a nested 5K control fixing
output cap 4096 in both arms.

Appendix~\ref{sec:prompts} specifies the controlled prompt differences; exact
UTF-8 files, output schemas, retry behavior, and provider-specific
settings are included in the supplementary software. The
intermediate domain in the two-step condition is an observable output. API calls allow
up to six attempts per instance. Exhausted, malformed, and policy-blocked outputs remain invalid
and are never rewritten as \texttt{000}.

\subsection{Metrics and Uncertainty}

These controlled conditions motivate evaluation at both the domain and code
levels. Let $y_i$ be the gold three-digit code and $\hat{y}_i$ the prediction. Exact
accuracy averages $\mathbb{1}[y_i=\hat{y}_i]$ over instances. Domain accuracy compares the first
digit. We additionally report macro and support-weighted $F_1$ over the 56 gold
labels. Invalid outputs remain in all denominators.

We decompose each prediction into:
\begin{itemize}[leftmargin=*,nosep]
    \item \textbf{exact correct};
    \item \textbf{within-domain error}: correct first digit, wrong code;
    \item \textbf{cross-domain error}: wrong first digit, including
    substantive-gold/\texttt{000} predictions; or
    \item \textbf{invalid output}: no legal CMP code was parsed.
\end{itemize}
This decomposition links the aggregate metrics to the paired prompt
contrasts. We use \emph{parent-domain selection} and \emph{within-domain code selection}
as operational descriptions of these observable task components. The oracle condition
both supplies the gold domain and restricts the candidate set to that domain's
children. We therefore interpret the flat--oracle contrast as a
privileged-information diagnostic of the performance recovered when the
correct parent and a restricted candidate set are provided. The full--restricted
comparison holds supplied domain information fixed and varies the candidate block.

With these outcomes defined, we construct 95\% confidence intervals for
accuracy and paired condition differences using 2,000 bootstrap replicates
clustered by manifesto. Clustering preserves within-document dependence. Since the
150K data are code-stratified, intervals
quantify uncertainty across manifestos represented in this benchmark rather
than sampling uncertainty for a naturally distributed corpus.
We also report deterministic lower anchors under the fixed benchmark design:
uniform chance over the 56 substantive labels, the optimal global constant
predictor under the benchmark label prior, and an oracle-domain majority
predictor that chooses the most frequent child after receiving the gold
domain. The majority predictor uses privileged information and is compared only
with the oracle-domain condition.

\section{Results}

\subsection{Cost-Efficient Models on 150K}

\begin{table*}[t]
\centering
\small
\begin{tabular}{lrrrrr}
\toprule
Model & Exact acc. & Domain acc. & Macro $F_1$ & Weighted $F_1$ & Invalid \\
\midrule
GPT-5.4-nano & 0.343 [0.335, 0.351] & 0.536 [0.529, 0.544] & 0.302 & 0.345 & 0.000 \\
Gemini 2.5 Flash Lite & 0.319 [0.309, 0.329] & 0.570 [0.557, 0.581] & 0.263 & 0.321 & 0.024 \\
Qwen Flash & 0.307 [0.299, 0.314] & 0.548 [0.541, 0.555] & 0.256 & 0.311 & 0.001 \\
Mistral Small & 0.211 [0.205, 0.216] & 0.384 [0.376, 0.391] & 0.164 & 0.205 & 0.005 \\
\midrule
Uniform chance (56 labels) & 0.018 & -- & -- & -- & -- \\
Optimal global constant & 0.033 & -- & -- & -- & -- \\
\bottomrule
\end{tabular}
\caption{Flat-prompt performance on the 150K benchmark. Accuracy intervals are 95\% manifesto-cluster bootstrap intervals. Invalid outputs remain in the denominator. Reference rows are deterministic anchors under the fixed benchmark label space and sample prior, so intervals and classification metrics do not apply.}
\label{tab:baseline-overall}
\end{table*}

We first establish the 150K performance profile for the four cost-efficient
models; Table~\ref{tab:configurations} compares all eight models on matched 30K inputs.
Table~\ref{tab:baseline-overall} shows that model rankings depend on
granularity. Among these four models, GPT-5.4-nano has the highest exact accuracy
(34.3\%) and weighted $F_1$ (34.5\%). Gemini is lower on exact coding (31.9\%)
but highest on domain accuracy (57.0\%). Qwen is close to Gemini, with 30.7\%
exact and 54.8\% domain accuracy. Mistral is substantially lower at both levels
(21.1\% and 38.4\%). Macro $F_1$ ranges from 16.4\% to 30.2\%, showing that
performance remains weak across labels even under code-stratified sampling.
For scale, uniform chance over the 56 gold labels is 1.8\%, and the best
constant prediction under the fixed benchmark prior reaches 3.3\%. Thus even
the weakest model substantially exceeds label-space chance and the benchmark's
optimal input-independent prediction, although all models remain far from
reliable exact-code reproduction.

Gemini has the largest invalid rate (2.43\%); GPT produces only two invalid
outputs in 150,000 evaluated occurrences. Because invalids remain in the
denominator, they cannot inflate model scores.

\subsection{Supervised Reference}

\begin{table}[t]
\centering
\scriptsize
\setlength{\tabcolsep}{2pt}
\begin{tabular}{lcccc}
\toprule
Model & Exact & Domain & Macro $F_1$ & Wtd. $F_1$ \\
\midrule
ManifestoBERTa & \shortstack{0.519\\{[0.498,0.538]}} & \shortstack{0.679\\{[0.663,0.692]}} & 0.470 & 0.511 \\
GPT-5.4-nano & \shortstack{0.337\\{[0.320,0.355]}} & \shortstack{0.534\\{[0.519,0.550]}} & 0.295 & 0.336 \\
Gemini 2.5 FL & \shortstack{0.311\\{[0.291,0.333]}} & \shortstack{0.566\\{[0.534,0.595]}} & 0.255 & 0.313 \\
Qwen Flash & \shortstack{0.306\\{[0.290,0.322]}} & \shortstack{0.553\\{[0.540,0.566]}} & 0.256 & 0.309 \\
Mistral Small & \shortstack{0.215\\{[0.205,0.225]}} & \shortstack{0.399\\{[0.384,0.412]}} & 0.167 & 0.209 \\
\bottomrule
\end{tabular}
\caption{Paired performance after excluding ManifestoBERTa 2023a corpus documents and normalized exact-text matches. Bracketed 95\% accuracy intervals use 2,000 manifesto-cluster bootstrap replicates.}
\label{tab:manifestoberta-no-overlap}
\end{table}

The main benchmark establishes instruction-only performance; the supervised
reference asks how much task-specific learning changes that picture. On the
screened subset, ManifestoBERTa reaches 51.9\% exact accuracy and
67.9\% domain accuracy, with macro and weighted $F_1$ of 47.0\% and 51.1\%
(Table~\ref{tab:manifestoberta-no-overlap}).
Among the four cost-efficient models on these rows, GPT's 33.7\% is the highest exact score;
ManifestoBERTa is 18.1 points higher (95\% paired cluster-bootstrap CI:
16.9--19.3). Its exact advantage ranges from 18.1 points over GPT to 30.4
points over Mistral, and every paired interval excludes zero. Its domain
advantage ranges from 11.3 to 28.0 points. Task-specific supervision therefore
provides a substantial external anchor for the instruction-only scores. It does not solve the task: 48.1\% of exact labels remain wrong, including 16.0\%
within-domain errors.

\subsection{Where Flat Predictions Fail}

The aggregate gap does not by itself reveal where errors occur.
We therefore decompose the four cost-efficient models' 150K outcomes. Cross-domain errors account
for 40.6\% of Gemini predictions, 45.2\% of Qwen predictions, 46.4\% of GPT
predictions, and 61.2\% of Mistral predictions. Within-domain errors account
for a further 17.3--25.0\%. The within-domain share is exactly the gain from exact
to domain accuracy: it captures cases where a model reaches the correct branch
but cannot reproduce the codebook-defined leaf label.

\begin{figure*}[t]
    \centering
    \includegraphics[width=\textwidth]{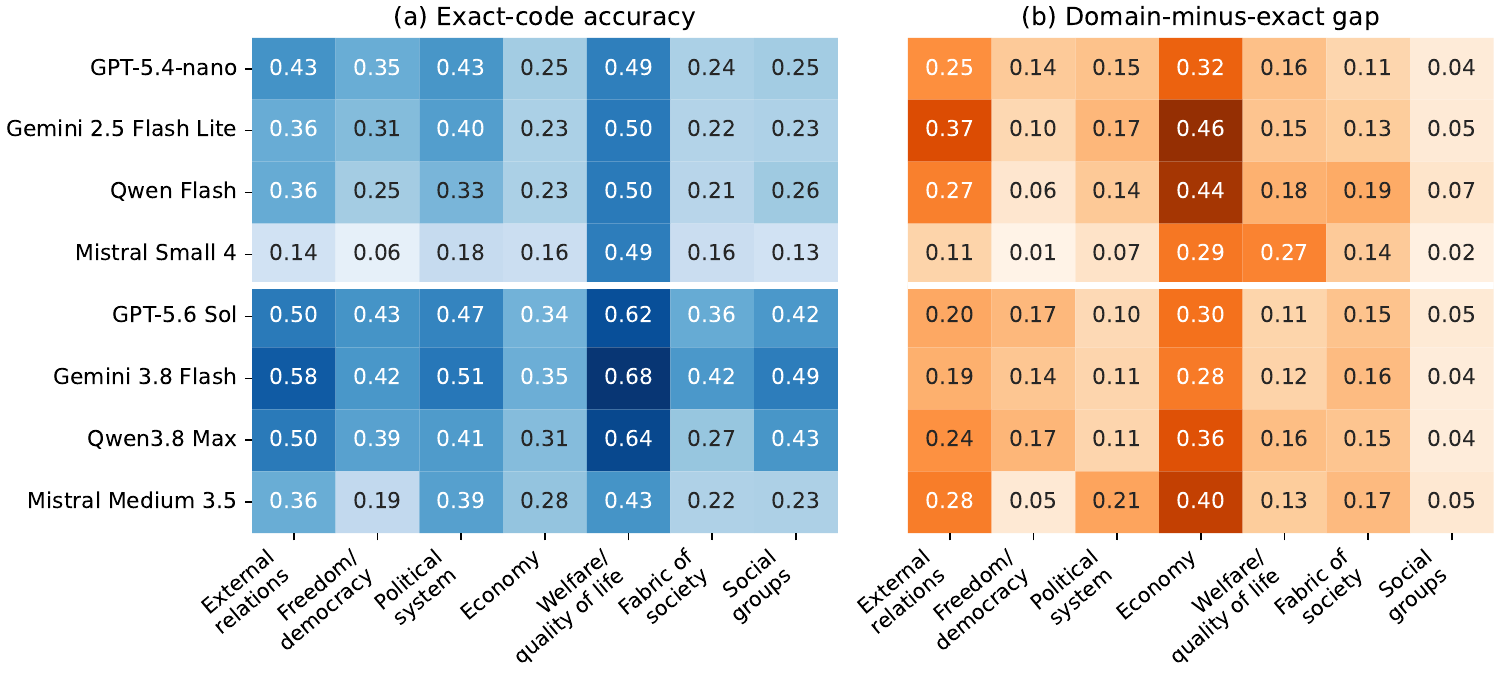}
    \caption{Domain-level results on the same 30K, with four cost-efficient
    models above four stronger models. Economy has the largest
    domain--code gap for every model (28--46 points).}
    \label{fig:domains}
\end{figure*}

Figure~\ref{fig:domains} extends the domain analysis to all eight models on
the same 30K. Welfare and Quality of Life has the highest exact accuracy for
every model (43.0--68.2\%). Economy has the largest domain--code gap in every
model (28.1--46.4 points), including the four stronger models. Many economic
labels therefore remain difficult after the correct topic has been recognized.

The domain-level pattern also appears in specific code pairs. On the paired
30K, the most frequent confusion for all eight models is 416 (Anti-Growth
Economy: Positive) predicted as 501 (Environmental Protection). For GPT,
Gemini and Qwen, stronger models increase overall accuracy while assigning
60.2--64.2\% of gold-416 instances to 501, compared with 37.7--50.6\% for
their base models. Frequent within-domain confusions include 402 (Incentives:
Positive) as 403 (Market Regulation), 503 (Equality: Positive) as 504 (Welfare
State Expansion), and 410 (Economic Growth: Positive) as 408 (Economic Goals)
(Appendix Table~\ref{tab:eight-model-confusions}). These comparisons locate
specific code boundaries requiring validation as model capability increases.

\subsection{Domain and Candidate-List Experiments}

\begin{table*}[t]
\centering
\small
\begin{tabular}{lrrrr}
\toprule
Model & Flat & Two-step & Gold/full & Gold/restricted \\
\midrule
GPT-5.4-nano & \shortstack[r]{0.341\\{\scriptsize [0.331, 0.351]}} & \shortstack[r]{0.353\\{\scriptsize [0.343, 0.362]}} & \shortstack[r]{0.417\\{\scriptsize [0.407, 0.426]}} & \shortstack[r]{0.554\\{\scriptsize [0.544, 0.563]}} \\
Gemini 2.5 Flash Lite & \shortstack[r]{0.319\\{\scriptsize [0.308, 0.330]}} & \shortstack[r]{0.340\\{\scriptsize [0.332, 0.348]}} & \shortstack[r]{0.499\\{\scriptsize [0.490, 0.509]}} & \shortstack[r]{0.532\\{\scriptsize [0.521, 0.542]}} \\
Qwen Flash & \shortstack[r]{0.307\\{\scriptsize [0.298, 0.317]}} & \shortstack[r]{0.315\\{\scriptsize [0.306, 0.324]}} & \shortstack[r]{0.427\\{\scriptsize [0.418, 0.437]}} & \shortstack[r]{0.533\\{\scriptsize [0.521, 0.544]}} \\
Mistral Small 4 & \shortstack[r]{0.209\\{\scriptsize [0.203, 0.216]}} & \shortstack[r]{0.296\\{\scriptsize [0.287, 0.304]}} & \shortstack[r]{0.459\\{\scriptsize [0.450, 0.468]}} & \shortstack[r]{0.502\\{\scriptsize [0.492, 0.513]}} \\
\midrule Oracle-domain majority & -- & -- & -- & 0.227 \\
\bottomrule
\end{tabular}
\caption{Exact-code accuracy on the fixed 30K paired sample, with 95\% manifesto-cluster bootstrap intervals. Both gold-domain conditions supply the correct parent; full-list uses all 56 titles and restricted-list uses its children. The oracle-majority reference is comparable to the restricted-list condition. Paired differences use unrounded scores. Settings appear in Appendix~\ref{app:inference-settings}; Appendix Table~\ref{tab:candidate-controls} reports candidate-list differences and fixed-budget 5K controls.}
\label{tab:mechanism-overall}
\end{table*}

The preceding decomposition is descriptive. The paired mechanism experiment
tests whether explicit parent-domain structure changes exact-code performance.
The fixed 30K flat scores closely reproduce the 150K ranking
(Table~\ref{tab:mechanism-overall}), confirming that the mechanism sample is a
useful diagnostic subset. Hierarchical two-step prompting improves exact
accuracy for all four cost-efficient models, with small gains for GPT (+1.1 points), Gemini
(+2.1), and Qwen (+0.8). Mistral gains 8.6 points. All four paired intervals
exclude zero, yet these gains remain far smaller than those in the gold-domain
condition.

Supplying the correct domain increases exact accuracy over flat prompting by
21.2 points for GPT, 21.3 for Gemini, 22.6 for Qwen, and 29.3 for Mistral
(Appendix Table~\ref{tab:oracle-effect}). Yet oracle accuracy remains between 50.2\% and
55.4\%. Thus supplying the gold domain and restricting the candidate set recovers a
large amount of performance but leaves nearly half of the codes incorrect.
An oracle-domain majority rule reaches only 22.7\% on the same preserved code
distribution (Table~\ref{tab:mechanism-overall}). The oracle LLM gains
therefore cannot be explained by always selecting the most frequent child of
the supplied domain.

Two-step domain/code consistency is at
least 99.9\% among valid outputs, and oracle constraints are respected in
95.6--99.98\% of predictions. Most violations are explicit \texttt{000}
outputs for sentences judged non-substantive despite the supplied domain.

\paragraph{Candidate-list restriction.}
With the gold domain and full list, exact accuracy is 41.67\%, 49.94\%,
42.69\% and 45.92\% for GPT, Gemini, Qwen and Mistral. Restriction adds
13.72, 3.26, 10.58 and 4.30 points, respectively; all paired intervals exclude
zero. Matched-cap 5K controls favor restriction in all four models
(Appendix Table~\ref{tab:candidate-controls}). These comparisons identify
candidate presentation as an additional contributor to the oracle gain.

\subsection{Eight-Model and Reasoning Comparisons}
\label{sec:configurations}
\begin{table*}[t]
\centering
\small
\begin{tabular}{lrrrr}
\toprule
Model/configuration & Exact & Domain & Macro $F_1$ & $\Delta$ exact [95\% CI] \\
\midrule
GPT-5.4-nano (base) & 34.14 & 53.43 & 29.81 & -- \\
GPT-5.6 Sol & 44.12 & 61.65 & 40.95 & +9.98 [+9.42, +10.55] \\
GPT-5.4-nano (high) & 34.93 & 53.84 & 31.19 & +0.79 [+0.44, +1.16] \\
Gemini 2.5 Flash Lite (base) & 31.94 & 56.63 & 25.97 & -- \\
Gemini 3.8 Flash & 48.35 & 65.40 & 44.50 & +16.41 [+15.50, +17.45] \\
Gemini 2.5 Flash Lite (high) & 34.58 & 57.43 & 30.54 & +2.64 [+1.90, +3.41] \\
Qwen Flash (base) & 30.72 & 54.64 & 25.68 & -- \\
Qwen3.8 Max & 41.29 & 61.86 & 35.25 & +10.58 [+10.04, +11.12] \\
Qwen Flash (high) & 33.11 & 54.06 & 30.42 & +2.39 [+1.93, +2.85] \\
Mistral Small 4 (base) & 20.95 & 38.06 & 16.27 & -- \\
Mistral Medium 3.5 & 30.88 & 53.10 & 26.94 & +9.94 [+9.23, +10.65] \\
Mistral Small 4 (high) & 34.57 & 56.90 & 30.00 & +13.62 [+12.92, +14.40] \\
\bottomrule
\end{tabular}
\caption{Flat-prompt configurations on the same 30K: scores in percent; differences in percentage points relative to each family's base flat row. Output caps are 256 for base flat (GPT permits retry expansion) and 4096 for stronger/high configurations. Appendix Table~\ref{tab:matched-configurations} reports configuration comparisons at a fixed cap of 4096. Intervals are paired manifesto-cluster bootstrap intervals.}
\label{tab:configurations}
\end{table*}

Stronger models improve exact accuracy in all four families, reaching
30.88--48.35\% (Table~\ref{tab:configurations}). High-reasoning gains over the
benchmark flat references are 0.79, 2.64, 2.39 and 13.62 points for GPT,
Gemini, Qwen and Mistral. Mistral Small-high reaches 34.57\%, compared with
30.88\% for the tested Medium configuration. Model tier and reasoning settings
therefore offer distinct routes to improving code assignment.

With the output cap fixed at 4096 tokens, higher reasoning increases exact
accuracy by 1.36, 2.20 and 13.62 points for Gemini, Qwen and Mistral,
respectively (Appendix Table~\ref{tab:matched-configurations}). For Qwen,
the exact-accuracy gain (95\% CI: 1.79--2.62) accompanies a domain-accuracy decrease of 0.94 points
(95\% CI: \textminus1.43 to \textminus0.46; Appendix Table~\ref{tab:matched-configurations}).
The two evaluation levels give opposite directions for this configuration change.

Across six paired 30K temperature comparisons, raising temperature from 0 to 0.7 changes
exact accuracy by \textminus0.92 to +0.05 points, while 23.59--35.51\% of
predictions change (Appendix Table~\ref{tab:temperature}). Five accuracy-difference
intervals include zero; Mistral at low reasoning decreases by 0.92 points
(95\% CI: \textminus1.25 to \textminus0.59). Score stability can coexist with
substantial instance-level variation.

\subsection{Duplicate-Annotation Sensitivity}

We repeat the baseline using one representative per unambiguous input group
and excluding 161 duplicated-input groups with conflicting corpus labels
($n=149{,}052$). Exact accuracy changes by less than 0.1 points for any model,
ruling out duplicate annotations as the source of the headline results.

\subsection{Descriptive Language Coverage}

The 32 detected document languages include 120,177 non-English instances
(80.1\%), and no language exceeds 19.9\% of the sample. Pooled non-English
exact accuracy is 0.1--1.3 points higher than English for GPT, Gemini, and Qwen, and
3.0 points lower for Mistral; the exact--domain gap persists in Appendix
Table~\ref{tab:language-performance}. These descriptive comparisons characterize
performance across language groups whose country, label and manifesto
composition varies.
Appendix Table~\ref{tab:eight-model-languages} compares all eight models within
the same 32 language groups on 30K.

\paragraph{Order robustness.}
On a fixed code-stratified 5K subset, reversing candidate-code order within
each domain changed GPT-5.4-nano exact accuracy by 0.3 percentage points (95\%
manifesto-cluster bootstrap CI: $-0.5$ to 1.1), while 24.1\% of predictions
changed. Aggregate accuracy was stable under this deterministic perturbation,
alongside substantial instance-level variation.

\section{Discussion}

\paragraph{Broad recognition and exact-code reproduction.}
Models encode meaningful information about the hierarchy without reliably
reproducing its leaf codes. Domain scoring alone would overstate coding
reliability, while exact scoring alone would hide useful routing ability.
Both levels and their decomposition are therefore necessary.

\paragraph{Task decomposition and candidate presentation.}
Two-step prompting enforces consistency but usually yields little gain, so
procedural compliance does not resolve codebook-defined boundaries. Even with
gold-domain information and a restricted candidate set, 44.6--49.8\% of predictions remain
incorrect. This differs from structured prediction, where constraints can
revise separately scored choices jointly \citep{pauk2026structured}. The
full-list ablation identifies candidate restriction as a source of the
oracle gain. Model and reasoning comparisons provide further configuration
choices for corpus-scale annotation.

\paragraph{Supervision and human agreement.}
ManifestoBERTa substantially outperforms the four cost-efficient instruction-only LLMs on
identical, screened inputs, yet still misses nearly half of exact codes and
retains a sizable exact--domain gap. The supervised and gold-domain comparisons
both identify fine-grained code selection as a persistent challenge.
\citet{mikhaylov2012coder} report median coder--master Cohen's $\kappa$ of
.43/.54 for UK/New Zealand instructional excerpts. Recomputing their released
labels with the original coder exclusions yields mean exact agreement of
47.44\%/59.95\% across 107/72 units and 17/12 coders. These historical results
provide external human-agreement context; agreement on our multilingual
sample requires its own multi-coder study.

\paragraph{Fixed coding schemes expose a complementary alignment setting.}
Taxonomy-refinement work improves classification through LLM-guided changes
to the hierarchy \citep{golde2026taxmorph}. CMP fixes code identities
and boundaries to support longitudinal and cross-national measurement;
changing them would change the target. Recurring confusions therefore diagnose
model--scheme misalignment at fixed category boundaries.

\paragraph{Multilingual scope and deployment.}
Four-fifths of instances are non-English, and language-level results extend
the evaluation across the benchmark's diverse document groups.
The results support using these systems to route statements and prioritize
human review, with final CMP codes validated by analysts. Such workflows
should retain outputs, prompt versions, and human corrections, and verify gains
in exact-code accuracy.

\section{Conclusion}

\bench{} tests exact reproduction of a 32-language hierarchical coding scheme.
On the 150,000-instance benchmark, cost-efficient models attain 21--34\% exact accuracy,
well below domain accuracy. A paired 30K study finds small two-step gains but
large, incomplete gold-domain gains; a training-screened ManifestoBERTa
reference reaches 51.9\%. Stronger models improve exact accuracy in all four families, reasoning
gains vary by model, and temperature changes can preserve aggregate scores
while changing individual predictions. These results
motivate reporting both hierarchy levels and complementing taxonomy adaptation
with fixed-scheme evaluation.

\section*{Limitations}
\label{sec:limitations}

These conclusions are bounded by several design choices. The benchmark fixes
one concise, provider-independent, sentence-only prompt.
Prompt and answer format can affect model behavior
\citep{labat2026polyglots}. The within-domain candidate-order reversal is a
single deterministic perturbation; broader prompt sensitivity remains to be
evaluated. We
also do not test retrieval, full-handbook conditioning,
surrounding context, constrained inference, model aggregation, or broader
supervised adaptation. ManifestoBERTa is one frozen supervised reference
for comparison.
Each temperature arm has one run; repeated generations would further
characterize variability.

A second set of limitations concerns data and generalization. Code
stratification changes the natural CMP label distribution. Language-level
comparisons are descriptive because language, country, manifesto, label mix,
and provider behavior are confounded. CMP reference labels are produced by
trained human coders and are imperfect \citep{mikhaylov2012coder}; removing
conflicting duplicate groups changes exact accuracy by less than 0.1 points, but
this check cannot replace fresh multi-coder adjudication. Public manifestos and
codebook material may also occur in model pretraining. Finally, our discussion
of medical, legal, regulatory, and scientific schemes is a methodological
generalization from CMP; we do not evaluate those domains empirically.

\section*{Ethical Considerations}

Election manifestos are public political documents, but automated coding can
create misleading claims about parties if model errors are ignored. We
recommend human validation for substantive use and reporting performance by
label and domain. The benchmark should not be used to infer individual-level
political attributes.

MPDS redistribution is governed by the Manifesto Project's terms, which do not
permit redistribution without written authorization. The
supplementary software therefore provides sampling and reconstruction scripts,
prompts, evaluation code, aggregate results, and audit summaries, but no
manifesto text, row-level labels, source-linked identifiers, or prediction
records. Researchers obtain MPDS 2025a and the Manifesto Corpus texts separately and reconstruct the
benchmark locally. The supplementary material includes dataset documentation
and an analysis manifest following dataset-documentation principles
\citep{gebru2021datasheets}.

Generative AI tools were used for language editing, drafting, and coding assistance.
The authors verified the final text, citations, analyses, and reported results.

\appendix
\raggedbottom

\section{Additional Experimental Results}
\label{app:mechanism-results}

\noindent\begin{minipage}{\columnwidth}
\centering
\small
\begin{tabular}{lr}
\toprule
Model & Oracle--flat (pp) \\
\midrule
GPT-5.4-nano & 21.2 [20.6, 21.9] \\
Gemini 2.5 Flash Lite & 21.3 [20.4, 22.3] \\
Qwen Flash & 22.6 [21.9, 23.3] \\
Mistral Small & 29.3 [28.5, 30.1] \\
\bottomrule
\end{tabular}
\captionof{table}{Paired oracle--flat difference in exact-code accuracy. Intervals use 2,000 manifesto-cluster bootstrap replicates.}
\label{tab:oracle-effect}
\end{minipage}

\begin{table*}[t]
\centering
\small
\begin{tabular}{lrr}
\toprule
Model & 30K difference [95\% CI] & Matched 5K difference [95\% CI] \\
\midrule
GPT-5.4-nano & +13.72 [+13.14, +14.28] & +14.36 [+13.05, +15.59] \\
Gemini 2.5 Flash Lite & +3.26 [+2.80, +3.72] & +3.02 [+2.01, +4.05] \\
Qwen Flash & +10.58 [+9.99, +11.16] & +11.00 [+9.71, +12.20] \\
Mistral Small 4 & +4.30 [+3.62, +4.96] & +3.92 [+2.73, +5.10] \\
\bottomrule
\end{tabular}
\caption{Restricted minus full-list exact agreement in percentage points. Both conditions supply the gold domain. The 30K settings appear in Appendix~\ref{app:inference-settings}; the 5K control fixes cap 4096 for both arms. All intervals are paired manifesto-cluster bootstrap intervals.}
\label{tab:candidate-controls}
\end{table*}

\begin{table*}[t]
\centering
\small
\begin{tabular}{llrrr}
\toprule
Base model & Compared configuration & Low exact & $\Delta$ exact [95\% CI] & $\Delta$ domain [95\% CI] \\
\midrule
Gemini 2.5 Flash Lite & Stronger & 33.22 & +15.14 [+14.49, +15.76] & +7.99 [+7.25, +8.77] \\
Gemini 2.5 Flash Lite & High reasoning & 33.22 & +1.36 [+0.91, +1.81] & +0.02 [-0.40, +0.45] \\
Qwen Flash & Stronger & 30.90 & +10.39 [+9.84, +10.98] & +6.86 [+6.28, +7.45] \\
Qwen Flash & High reasoning & 30.90 & +2.20 [+1.79, +2.62] & -0.94 [-1.43, -0.46] \\
Mistral Small 4 & Stronger & 20.94 & +9.94 [+9.22, +10.66] & +14.60 [+13.55, +15.67] \\
Mistral Small 4 & High reasoning & 20.94 & +13.62 [+12.94, +14.38] & +18.41 [+17.47, +19.46] \\
\bottomrule
\end{tabular}
\caption{Output-budget-controlled configuration comparisons on 30K. Each difference is relative to low-reasoning flat inference for the same base model; all conditions use output cap 4096. Exact API settings appear in Appendix~\ref{app:inference-settings}. Differences are percentage points with paired manifesto-cluster intervals.}
\label{tab:matched-configurations}
\end{table*}

\begin{table*}[t]
\centering
\small
\begin{tabular}{llrrrr}
\toprule
Model & Reasoning & $T=0$ & $T=0.7$ & $\Delta$ exact [95\% CI] & Changed (\%) \\
\midrule
Gemini 2.5 Flash Lite & Low & 33.22 & 32.88 & -0.33 [-0.70, +0.03] & 25.06 \\
Gemini 2.5 Flash Lite & High & 34.58 & 34.44 & -0.13 [-0.50, +0.22] & 27.83 \\
Qwen Flash & Low & 30.90 & 30.91 & +0.01 [-0.32, +0.33] & 23.59 \\
Qwen Flash & High & 33.11 & 33.06 & -0.05 [-0.39, +0.30] & 25.21 \\
Mistral Small 4 & Low & 20.94 & 20.02 & -0.92 [-1.25, -0.59] & 35.31 \\
Mistral Small 4 & High & 34.57 & 34.62 & +0.05 [-0.39, +0.49] & 35.51 \\
\bottomrule
\end{tabular}
\caption{Temperature sensitivity on 30K at fixed model, prompt, reasoning setting and output cap (4096). Exact agreement is in percent and differences in percentage points. Changed counts all differing parsed predictions, with invalid responses pooled as INVALID. Zero-temperature high references reuse the corresponding reasoning runs. Each arm is one run; intervals resample manifestos.}
\label{tab:temperature}
\end{table*}

\begin{table*}[t]
\centering
\small
\begin{tabular}{lrrrr}
\toprule
Model & Gold & Predicted & Count & Gold-class rate (\%) \\
\midrule
GPT-5.4-nano & 416 & 501 & 506 & 50.6 \\
GPT-5.4-nano & 305 & 000 & 284 & 38.9 \\
GPT-5.4-nano & 601 & 000 & 173 & 18.8 \\
GPT-5.6 Sol & 416 & 501 & 608 & 60.8 \\
GPT-5.6 Sol & 305 & 000 & 227 & 31.1 \\
GPT-5.6 Sol & 601 & 608 & 191 & 20.8 \\
Gemini 2.5 Flash Lite & 416 & 501 & 455 & 45.5 \\
Gemini 2.5 Flash Lite & 410 & 408 & 311 & 34.3 \\
Gemini 2.5 Flash Lite & 701 & 408 & 277 & 27.7 \\
Gemini 3.8 Flash & 416 & 501 & 642 & 64.2 \\
Gemini 3.8 Flash & 607 & 705 & 214 & 37.6 \\
Gemini 3.8 Flash & 601 & 608 & 213 & 23.2 \\
Qwen Flash & 416 & 501 & 377 & 37.7 \\
Qwen Flash & 402 & 403 & 265 & 26.5 \\
Qwen Flash & 305 & 000 & 241 & 33.0 \\
Qwen3.8 Max & 416 & 501 & 602 & 60.2 \\
Qwen3.8 Max & 503 & 504 & 211 & 21.1 \\
Qwen3.8 Max & 305 & 000 & 155 & 21.2 \\
Mistral Small 4 & 416 & 501 & 402 & 40.2 \\
Mistral Small 4 & 701 & 504 & 375 & 37.5 \\
Mistral Small 4 & 305 & 000 & 265 & 36.3 \\
Mistral Medium 3.5 & 416 & 501 & 397 & 39.7 \\
Mistral Medium 3.5 & 202 & 303 & 256 & 25.6 \\
Mistral Medium 3.5 & 503 & 504 & 188 & 18.8 \\
\bottomrule
\end{tabular}
\caption{Three most frequent valid misclassification pairs per model on the same 30K. Gold-class rate divides the pair count by all instances with that gold code. Invalid outputs remain in accuracy denominators.}
\label{tab:eight-model-confusions}
\end{table*}

\section{Reproducibility Details}
\label{app:reproducibility}

\subsection{Fixed Samples}

The main sample is generated locally with seed 20260629. The mechanism sample
is generated locally with seed 20260713. Supplementary conditions use the
same fixed 30K sample. Predictions are paired by sentence identifier.

\subsection{Inference Settings}
\label{app:inference-settings}

The 150K baseline contains 600,000 model--instance outcomes. The four-model
two-step and restricted-oracle experiments add 240,000 outcomes on 30K.
Candidate-list, stronger-model, reasoning and temperature comparisons also
use this fixed 30K; local ManifestoBERTa inference uses the screened subset.

\begin{center}
\scriptsize
\begin{tabular}{lcc}
\toprule
Model & Reasoning & Cap: flat/mechanism \\
\midrule
GPT-5.4-nano & low & 256 / 512 \\
Gemini 2.5 Flash Lite & budget 512 & 256 / 1024 \\
Qwen Flash & budget 256 & 256 / 512 \\
Mistral Small 4 & none & 256 / 512 \\
\bottomrule
\end{tabular}
\end{center}
\noindent{\small\textit{Provider-specific settings. Mechanism conditions use
larger output caps for structured responses and the same reasoning settings
as the flat conditions.}}

\paragraph{Model and configuration comparisons.}
Full-list, stronger-model and high-reasoning conditions use output cap 4096.
The 150K, two-step and oracle runs request \path{gpt-5.4-nano},
\path{gemini-2.5-flash-lite}, \path{qwen-flash} and \path{mistral-small-latest}.
Base identifiers for the supplementary configuration comparisons are
\path{gpt-5.4-nano-2026-03-17}, \path{gemini-2.5-flash-lite},
\path{qwen-flash-2025-07-28} and \path{mistral-small-2603}.
Stronger identifiers are \path{gpt-5.6-sol}, \path{gemini-3.8-flash},
\path{qwen3.8-max-0902} and \path{mistral-medium-3-5}.
Stronger-model reasoning is low for GPT and Gemini, budget 256 for Qwen,
and none for Mistral. High base-model reasoning is high for GPT/Mistral and
budget 2048 for Gemini/Qwen. GPT verbosity is low; other providers omit this
field. Temperature is zero except in the 0.7 arms; GPT omits temperature.
For Gemini, Qwen and Mistral, the fixed-budget reasoning comparison uses the
same flat prompt and cap 4096 at low and high reasoning. Temperature comparisons
vary 0 versus 0.7 within each of these settings. The temperature-zero conditions
are shared across the reasoning and temperature analyses. Each temperature
arm has one run. The software records requested model identifiers and settings.

\subsection{Prompt and Output Audit}
\label{app:prompt-audit}

Validation confirms that all 56 candidate titles match the shared flat
baseline exactly and that each oracle candidate is placed under the correct
domain.

\subsection{Analysis Commands}

Without non-redistributed manifesto text and row-level records, reviewers run
\texttt{python verify\_release.py} to check the released aggregate results.
After reconstructing those inputs and outputs, \texttt{python reproduce.py} rebuilds all reported metrics,
intervals, tables, and figures under \texttt{outputs/}.
This command also generates the supplementary 30K tables and paired intervals
from saved predictions, without API calls.

\section{Language Coverage}

\begin{table*}[t]
\centering
\scriptsize
\setlength{\tabcolsep}{3.2pt}
\begin{tabular}{lrrrcccc}
\toprule
Language & Countries & Manifestos & Sentences & GPT & Gemini & Qwen & Mistral \\
\midrule
English & 7 & 75 & 29,823 & 33.5/52.6 & 30.9/55.9 & 30.6/53.8 & 23.4/43.0 \\
Spanish & 7 & 66 & 19,152 & 36.5/54.6 & 33.6/55.2 & 33.1/54.9 & 22.3/39.5 \\
Dutch & 2 & 24 & 14,611 & 32.4/51.6 & 29.4/54.3 & 29.0/52.7 & 18.8/36.6 \\
German & 3 & 30 & 13,496 & 38.2/58.1 & 33.6/60.5 & 33.8/58.5 & 21.1/37.7 \\
Portuguese & 2 & 27 & 10,490 & 35.2/55.1 & 33.9/60.5 & 31.6/57.8 & 22.5/42.2 \\
French & 4 & 17 & 7,621 & 29.5/46.7 & 25.1/43.4 & 28.0/49.6 & 18.6/34.4 \\
Croatian & 4 & 44 & 6,712 & 35.4/55.5 & 34.7/62.7 & 31.7/57.7 & 21.7/38.1 \\
Hebrew & 1 & 31 & 5,608 & 35.8/54.8 & 34.5/59.3 & 31.0/55.7 & 20.6/36.5 \\
Greek & 1 & 13 & 4,336 & 19.9/45.1 & 22.8/57.6 & 18.4/51.5 & 13.8/33.1 \\
Catalan & 1 & 9 & 4,266 & 38.7/55.5 & 35.8/57.0 & 32.0/53.3 & 19.2/34.0 \\
Hungarian & 1 & 10 & 4,126 & 32.7/50.4 & 31.7/54.7 & 28.8/49.1 & 16.9/29.1 \\
Turkish & 1 & 6 & 3,816 & 35.0/56.4 & 31.8/61.2 & 30.1/57.8 & 20.2/38.0 \\
Italian & 2 & 12 & 3,298 & 29.7/47.8 & 27.7/51.5 & 26.1/46.6 & 17.6/30.2 \\
Swedish & 2 & 17 & 2,317 & 38.7/57.8 & 36.0/59.9 & 34.5/59.0 & 25.8/41.1 \\
Lithuanian & 1 & 8 & 2,184 & 37.4/58.3 & 39.6/64.8 & 33.8/60.3 & 24.9/43.1 \\
Finnish & 1 & 8 & 2,033 & 35.9/54.8 & 30.2/51.4 & 30.6/54.5 & 22.7/39.0 \\
Danish & 1 & 10 & 1,900 & 27.6/45.9 & 27.3/51.6 & 27.4/50.7 & 21.1/40.6 \\
Polish & 1 & 7 & 1,723 & 33.0/50.4 & 32.4/59.3 & 28.8/52.8 & 21.6/37.6 \\
Slovenian & 1 & 9 & 1,688 & 30.6/50.8 & 30.1/57.2 & 27.0/53.6 & 18.7/35.2 \\
Korean & 1 & 5 & 1,671 & 36.3/56.0 & 33.6/59.8 & 31.5/57.8 & 15.7/27.0 \\
Japanese & 1 & 8 & 1,533 & 34.9/59.3 & 30.1/64.4 & 32.4/62.8 & 18.4/33.5 \\
Georgian & 1 & 8 & 1,410 & 36.2/57.2 & 35.6/64.8 & 30.7/58.9 & 22.9/41.5 \\
Estonian & 1 & 5 & 1,117 & 41.1/60.3 & 39.7/66.4 & 32.7/59.5 & 21.2/39.1 \\
Serbian & 3 & 16 & 1,060 & 37.6/56.4 & 37.5/61.9 & 31.2/52.9 & 21.8/32.6 \\
Maltese & 1 & 2 & 957 & 34.7/57.2 & 34.6/63.0 & 27.7/55.6 & 23.2/41.0 \\
Armenian & 1 & 7 & 927 & 39.8/61.3 & 39.3/68.8 & 34.6/63.9 & 21.4/36.9 \\
Romanian & 1 & 5 & 684 & 38.5/58.5 & 38.9/64.8 & 37.7/59.9 & 22.5/38.7 \\
Icelandic & 1 & 8 & 444 & 38.3/58.3 & 33.1/56.3 & 28.8/53.8 & 19.6/36.7 \\
Galician & 1 & 2 & 362 & 43.6/60.8 & 35.1/58.0 & 31.8/49.4 & 19.9/30.7 \\
Latvian & 1 & 15 & 306 & 33.0/54.9 & 35.6/61.8 & 28.1/52.3 & 16.3/31.0 \\
Ukrainian & 1 & 6 & 186 & 43.5/65.1 & 38.7/71.0 & 34.9/66.7 & 21.5/38.2 \\
Albanian & 1 & 3 & 143 & 46.2/65.0 & 50.3/73.4 & 37.8/58.7 & 27.3/39.2 \\
\bottomrule
\end{tabular}
\caption{Descriptive coverage and flat-prompt performance by document language. Model cells give exact/domain accuracy in percent. Language is assigned once per full available original-language document; groups are descriptive, not causal. Country counts are within-language counts and are not additive because multilingual countries may appear in multiple rows. Row sizes also reflect source-document length and code-stratified sampling.}
\label{tab:language-performance}
\end{table*}

\begin{table*}[t]
\centering
\scriptsize
\setlength{\tabcolsep}{2pt}
\begin{tabular}{lrrrrrrrrr}
\toprule
Language & $n$ & GPT-N & GPT-S & Gem-L & Gem-F & Q-F & Q-M & M-S & M-M \\
\midrule
English & 5,983 & 33.3/53.0 & 42.3/60.1 & 30.1/54.9 & 45.2/61.9 & 29.2/53.5 & 38.9/59.9 & 22.9/42.7 & 30.0/52.6 \\
Spanish & 3,908 & 36.5/53.6 & 45.2/60.6 & 33.9/54.8 & 49.0/64.4 & 34.0/54.8 & 43.2/61.4 & 22.4/39.5 & 32.7/52.9 \\
Dutch & 2,898 & 34.0/51.2 & 43.0/59.6 & 30.6/54.7 & 46.1/62.2 & 30.2/52.1 & 39.4/60.4 & 20.3/36.3 & 31.4/52.8 \\
German & 2,707 & 38.5/58.0 & 48.1/65.1 & 33.1/59.8 & 51.9/68.1 & 34.4/58.0 & 44.5/64.3 & 20.5/37.7 & 33.8/55.5 \\
Portuguese & 2,130 & 33.8/54.5 & 43.7/61.5 & 32.5/58.5 & 50.1/67.0 & 32.1/57.9 & 42.7/63.2 & 22.1/41.2 & 31.2/53.1 \\
French & 1,525 & 29.6/46.4 & 39.9/55.7 & 25.4/43.1 & 44.5/62.5 & 28.1/50.2 & 38.3/58.4 & 19.5/35.2 & 25.7/46.4 \\
Croatian & 1,338 & 35.9/56.7 & 46.5/63.9 & 36.3/62.7 & 49.3/68.5 & 32.8/58.6 & 43.8/65.5 & 21.1/37.9 & 32.6/55.2 \\
Hebrew & 1,087 & 33.8/53.0 & 44.5/62.5 & 33.4/57.4 & 49.5/66.9 & 27.8/52.7 & 41.0/60.1 & 18.1/34.2 & 31.2/53.4 \\
Greek & 892 & 17.8/43.3 & 27.5/55.3 & 22.9/58.2 & 32.6/62.6 & 17.7/48.9 & 29.7/61.0 & 14.6/33.2 & 23.0/54.5 \\
Hungarian & 806 & 33.6/52.5 & 46.2/64.4 & 32.9/56.7 & 52.1/69.0 & 28.7/49.8 & 42.4/62.7 & 19.2/31.0 & 26.8/49.1 \\
Catalan & 806 & 38.2/54.2 & 51.6/62.9 & 36.0/55.6 & 57.2/67.0 & 32.9/51.0 & 45.3/59.6 & 18.6/32.8 & 31.8/49.8 \\
Turkish & 752 & 35.0/56.4 & 47.6/70.3 & 31.1/61.3 & 52.4/72.6 & 31.8/60.0 & 42.8/68.1 & 19.8/37.9 & 31.4/56.9 \\
Italian & 652 & 30.4/48.3 & 39.7/55.2 & 29.8/52.9 & 43.4/58.3 & 26.2/46.9 & 35.3/52.9 & 15.6/27.6 & 29.9/49.4 \\
Swedish & 442 & 40.3/59.3 & 51.8/69.9 & 39.1/63.3 & 53.2/70.1 & 37.8/62.7 & 45.7/67.6 & 26.5/40.5 & 38.7/62.7 \\
Lithuanian & 432 & 35.2/59.3 & 43.8/64.1 & 35.9/65.0 & 49.3/70.4 & 31.7/61.6 & 47.0/69.9 & 22.7/44.0 & 28.9/52.8 \\
Danish & 412 & 26.9/44.7 & 38.3/56.3 & 29.4/55.6 & 46.4/64.3 & 27.4/51.2 & 36.2/54.4 & 21.4/41.5 & 28.2/51.5 \\
Finnish & 384 & 35.9/54.9 & 46.4/68.2 & 30.5/52.1 & 49.0/70.8 & 30.7/55.7 & 42.2/66.9 & 19.3/35.7 & 31.8/57.0 \\
Polish & 368 & 29.1/49.2 & 41.0/57.3 & 28.3/54.1 & 50.3/64.7 & 28.5/53.3 & 37.5/56.2 & 19.8/35.9 & 30.2/52.4 \\
Slovenian & 327 & 32.1/52.9 & 45.6/60.2 & 33.6/59.0 & 50.2/64.8 & 32.4/54.4 & 45.0/61.5 & 19.6/35.5 & 32.7/52.9 \\
Korean & 314 & 35.0/52.5 & 46.5/63.7 & 37.3/59.9 & 55.1/70.4 & 32.8/57.6 & 39.8/62.4 & 16.2/29.0 & 24.5/49.7 \\
Japanese & 295 & 32.9/58.3 & 45.8/65.1 & 26.8/62.4 & 51.9/68.8 & 30.5/62.7 & 41.4/63.7 & 19.0/31.5 & 29.5/58.0 \\
Georgian & 287 & 33.8/62.0 & 47.0/69.3 & 29.6/65.5 & 54.7/76.7 & 24.7/59.2 & 45.6/72.8 & 24.7/47.0 & 30.0/54.4 \\
Estonian & 228 & 41.2/58.8 & 57.5/74.6 & 42.1/68.0 & 60.5/75.4 & 34.2/61.0 & 52.2/71.1 & 19.7/32.0 & 34.6/55.7 \\
Maltese & 214 & 35.5/58.4 & 39.7/63.1 & 34.6/65.4 & 44.9/65.0 & 25.7/55.6 & 42.1/63.1 & 20.6/36.4 & 29.4/54.2 \\
Serbian & 201 & 39.8/59.7 & 53.2/71.1 & 38.8/62.2 & 54.7/72.1 & 30.3/53.2 & 49.3/68.7 & 21.9/32.3 & 34.8/57.2 \\
Armenian & 187 & 36.9/57.8 & 49.2/72.7 & 37.4/66.3 & 58.3/79.1 & 36.4/64.7 & 46.5/70.6 & 21.4/37.4 & 32.1/55.1 \\
Romanian & 141 & 41.8/61.7 & 48.2/62.4 & 41.8/64.5 & 56.0/69.5 & 41.8/64.5 & 48.9/68.8 & 24.8/36.2 & 44.0/61.7 \\
Icelandic & 90 & 35.6/55.6 & 51.1/65.6 & 38.9/57.8 & 54.4/70.0 & 30.0/54.4 & 45.6/64.4 & 30.0/46.7 & 33.3/53.3 \\
Galician & 74 & 41.9/59.5 & 51.4/63.5 & 33.8/52.7 & 59.5/68.9 & 32.4/47.3 & 44.6/56.8 & 14.9/23.0 & 35.1/48.6 \\
Latvian & 59 & 33.9/62.7 & 44.1/67.8 & 47.5/76.3 & 52.5/72.9 & 27.1/57.6 & 42.4/62.7 & 15.3/30.5 & 25.4/59.3 \\
Ukrainian & 37 & 35.1/51.4 & 51.4/73.0 & 43.2/70.3 & 56.8/75.7 & 29.7/59.5 & 43.2/70.3 & 18.9/37.8 & 37.8/64.9 \\
Albanian & 24 & 50.0/66.7 & 54.2/70.8 & 62.5/79.2 & 58.3/75.0 & 41.7/66.7 & 66.7/83.3 & 33.3/41.7 & 37.5/58.3 \\
\bottomrule
\end{tabular}
\caption{Eight-model flat-prompt comparison on the same 30K. Cells show exact/domain accuracy in percent. GPT-N/S: GPT-5.4-nano/GPT-5.6 Sol; Gem-L/F: Gemini 2.5 Flash Lite/Gemini 3.8 Flash; Q-F/M: Qwen Flash/Qwen3.8 Max; M-S/M: Mistral Small 4/Mistral Medium 3.5. Languages are assigned at document level. Group sizes reflect the fixed sample.}
\label{tab:eight-model-languages}
\end{table*}

Appendix Table~\ref{tab:language-performance} is descriptive. Languages are
identified once from each full available source manifesto rather than independently
from short quasi-sentences. The local row-level audit records each assignment,
while the supplementary software reports aggregate language coverage and
performance without redistributing source-linked identifiers.

\section{Prompt Specification}
\label{sec:prompts}

All conditions share the coder role, specificity and salience rules, key
distinctions, and eight examples. The supplementary software contains the
canonical UTF-8 prompts audited in Appendix~\ref{app:prompt-audit}.
The conditions are specified below:

\begin{description}[leftmargin=0pt,labelindent=0pt,itemsep=2pt,topsep=2pt]
\item[Flat codebook prompt.] The prompt defines a quasi-sentence as a single
political idea or policy position and lists all 56 canonical
three-digit code-title pairs, allows 000 for statements without political
content, and requires one three-digit code.
\item[Hierarchical two-step.] The prompt describes domains 0--7 and lists the
same 56 canonical code-title pairs grouped by domain. It requires JSON fields
\texttt{step1\_domain} and \texttt{step2\_code}.
\item[Oracle domain.] The prompt supplies the gold domain number and title and
lists only the canonical child code-title pairs legal in that domain. It
requires one code from that candidate set.
\item[Gold domain/full list.] The same oracle template supplies the gold
domain; the candidate block lists all 56 titles in canonical order.
\item[Configuration comparisons.] Model, reasoning and temperature
experiments reuse the flat prompt verbatim.
\end{description}

\paragraph{Runtime instantiation.}
For the oracle condition, the placeholders
\texttt{\{DOMAIN\}}, \texttt{\{DOMAIN\_NAME\}}, and
\texttt{\{DOMAIN\_CODES\}} are replaced before every API call. Consequently,
the model receives the numeric mapping for every legal child category, for
example ``501 Environmental Protection'' through ``507 Education Limitation''
when domain 5 is supplied. The quasi-sentence is appended after the fixed
instruction in every condition.

\paragraph{Output parsing.}
For flat and oracle responses, the parser extracts the first three-digit
sequence and accepts it only if it is a legal CMP main code. Two-step responses
are parsed using the two-field JSON schema above. Legal domain/code
inconsistencies and oracle candidate-set violations are retained and flagged
separately; malformed or missing codes are invalid rather than mapped to 000.
The supplementary prompt files are the authoritative verbatim specification;
this appendix presents their experimentally relevant structure in a readable
form.

\end{document}